\documentclass[11pt, a4paper]{article}

\usepackage[utf8]{inputenc}
\usepackage[margin=1in]{geometry} 
\usepackage{authblk}              
\usepackage{graphicx}             
\usepackage{booktabs}             
\usepackage{hyperref}             
\usepackage{xcolor}               
\usepackage{listings}             
\usepackage{amsmath}              
\usepackage{float}                
\usepackage{enumitem}
\usepackage{xcolor}

\definecolor{codegray}{rgb}{0.5,0.5,0.5}
\definecolor{codegreen}{rgb}{0,0.6,0}
\definecolor{codepurple}{rgb}{0.58,0,0.82}
\definecolor{backcolour}{rgb}{0.95,0.95,0.92}

\lstdefinelanguage{json}{
    basicstyle=\normalfont\ttfamily,
    numbers=left,
    numberstyle=\scriptsize,
    stepnumber=1,
    numbersep=8pt,
    showstringspaces=false,
    breaklines=true,
    frame=lines,
    string=[s]{"}{"},
    comment=[l]{//},
    morecomment=[s]{/*}{*/},
    stringstyle=\color{blue},
    literate=
     *{0}{{{\color{orange}0}}}{1}
      {1}{{{\color{orange}1}}}{1}
      {2}{{{\color{orange}2}}}{1}
      {3}{{{\color{orange}3}}}{1}
      {4}{{{\color{orange}4}}}{1}
      {5}{{{\color{orange}5}}}{1}
      {6}{{{\color{orange}6}}}{1}
      {7}{{{\color{orange}7}}}{1}
      {8}{{{\color{orange}8}}}{1}
      {9}{{{\color{orange}9}}}{1}
      {:}{{{\color{magenta}{:}}}}{1}
      {,}{{{\color{magenta}{,}}}}{1}
      {\{}{{{\color{cyan}{\{}}}}{1}
      {\}}{{{\color{cyan}{\}}}}}{1}
      {[}{{{\color{cyan}{[}}}}{1}
      {]}{{{\color{cyan}{]}}}}{1},
}

\lstdefinestyle{mystyle}{
    backgroundcolor=\color{backcolour},   
    commentstyle=\color{codegreen},
    keywordstyle=\color{magenta},
    numberstyle=\tiny\color{codegray},
    stringstyle=\color{codepurple},
    basicstyle=\ttfamily\footnotesize,
    breakatwhitespace=false,         
    breaklines=true,                 
    captionpos=b,                    
    keepspaces=true,                 
    numbers=left,                    
    numbersep=5pt,                  
    showspaces=false,                
    showstringspaces=false,
    showtabs=false,                  
    tabsize=2
}

\title{\textbf{Optimizing Small Language Models for NL2SQL via Chain-of-Thought Fine-Tuning}}

\author{Anshul Solanki, Koushik Chakraborty, Navneet Kamboj and Sanchit Latawa}

\affil{Google AI, Global Services Delivery}

\date{}

\begin{document}

\maketitle

\begin{abstract}
Translating Natural Language to SQL (NL2SQL) remains a critical bottleneck for democratization of data in enterprises. Although large language models (LLMs) such as Gemini 2.5 and other LLMs have demonstrated impressive zero-shot capabilities, their high inference costs limit deployment at scale. This paper explores the efficacy of fine-tuning both large and small language models on NL2SQL tasks.

Our research reveals a counter-intuitive scaling phenomenon. Fine-tuning large models (Gemini 2.5 Flash/Lite) on standard datasets yields negligible returns, often leading to overfitting on complex queries. In contrast, small models (Qwen) show significant gains. Fine-tuning improved the small model baseline from 36\% to 45\%, and further enriching the data set with explicit Chain-of-Thought (CoT) reasoning increased accuracy to 54.5\%(Fig 2). Although this is still lower than the accuracy of large models like Gemini 2.5 , it does serve the business goal of significant cost reduction, latency in inference time, and also meeting the business critical performance accuracy threshold.This paper demonstrates that transferring reasoning patterns enables compute-efficient smaller models to approach production-grade performance.
\end{abstract}

\section{Introduction}
As organizations increasingly rely on data-driven decision-making, the ability for non-technical users to query relational databases using natural language (NL2SQL) has become highly desirable.
Conventionally, deploying high-accuracy NL2SQL agents require utilizing state-of-the-art, massive parameter LLMs ($>$70B parameters) or specialized architectures~\cite{ratsql}. While effective, these models introduce significant latency and compute costs. We investigated whether fine-tuning could bridge the gap between resource-heavy models and smaller, compute-efficient models like Qwen~\cite{qwen}.

\section{Related Work: NL2SQL and Benchmarks}
\paragraph{NL2SQL and Benchmarks:}
The task of translating natural language to SQL has historically relied on encoder-decoder architectures trained on large-scale datasets. The release of the Spider dataset~\cite{spider} shifted the focus toward cross-domain generalization and complex nested queries. While early approaches utilized specialized architectures like RAT-SQL~\cite{ratsql}, recent state-of-the-art performance has been dominated by Large Language Models (LLMs) such as GPT-4 and Gemini using in-context learning~\cite{gemini}.

\paragraph{Chain-of-Thought Reasoning:}
Wei et al.~\cite{cot} demonstrated that prompting LLMs to generate intermediate reasoning steps significantly improves performance on complex logical tasks. This ``Chain-of-Thought'' (CoT) capability is typically emergent only in models larger than 100B parameters. Recent work on ``Knowledge Distillation''~\cite{distill} suggests that reasoning capabilities can be transferred to smaller student models. Our work extends this by explicitly fine-tuning small models on CoT traces, effectively distilling the reasoning process of larger models into a compute-efficient architecture.

\section{Methodology and Evaluation Framework}
To ensure rigorous benchmarking, we implemented a standardized evaluation pipeline during the research.

\subsection{Evaluation Metrics}
Instead of relying on text-based ``Exact Match'' metrics, which often penalize valid SQL variations, we implemented \textbf{Execution Accuracy}. We execute the generated SQL against a shadow database and compare the result set against the ground truth, a standard practice in modern NL2SQL research~\cite{spider}.

\subsection{Dataset \& Filtering Strategy: Spider-1.0}
We refined the Spider-1.0 dataset to focus on complex queries—specifically those involving advanced percentiles and formulas—alongside a representative 600-query benchmark. This curation was important as modern LLMs already demonstrate nearly perfect accuracy on basic queries, necessitating a focus on high-difficulty structures to drive model improvement.

\subsection{Complexity Scoring Methodology}
The training data was filtered based on a custom complexity scoring of the SQL structures. We devised a formula to score SQL queries by adding complexity points for specific clauses: \texttt{JOIN}, \texttt{GROUP BY}, \texttt{ORDER BY}, \texttt{HAVING}, and recursively scoring nested queries found in sub-queries or set operators. For example, a query containing one nested subquery in its \texttt{WHERE} clause earns a higher score as it triggers a level of recursion.

\subsection{Dataset Distributions}
To maximize learning density, we shifted from the source distribution toward a model optimized for "Hard" queries.

\paragraph{Initial Source Distribution (7,000 Queries):}
The original source distribution showed a heavy tilt toward medium-difficulty tasks, as detailed in Table \ref{tab:initial_dist}.

\begin{table}[h]
\centering
\small
\begin{tabular}{@{}llll@{}}
\toprule
\textbf{Difficulty} & \textbf{Query Count} & \textbf{Complexity Score Range} & \textbf{Median Score} \\ \midrule
Easy                & 2,100                & 0 (Min) -- 1 (Max)              & 0.0                   \\
Medium              & 2,800                & 1 (Min) -- 3 (Max)              & 2.0                   \\
Hard                & 2,100                & 3 (Min) -- 20 (Max)             & 4.0                   \\ \bottomrule
\end{tabular}
\caption{Complexity distribution of the source 7,000 query train\_dataset.}
\label{tab:initial_dist}
\end{table}

\paragraph{Optimized Training Distribution (5,500 Queries):}
Our optimized dataset (5,500 total queries) follows a refined distribution to minimize trivial queries. This curated data is partitioned into training and validation sets at a \textbf{5,000:500 ratio}.

\begin{itemize}[label=\textbullet, noitemsep]
    \item \textbf{Hard (40\%):} High-complexity nested logic.
    \item \textbf{Medium (50\%):} Standard relational operations.
    \item \textbf{Easy (10\%):} Essential baseline queries.
\end{itemize}

\subsection{Benchmark Configuration}
We applied the same algorithmic complexity distribution to the 600-query benchmark data to ensure evaluation results reflect real-world difficulty:
\begin{itemize}[label=--, noitemsep]
    \item \textbf{Benchmark Total Count:} 600
    \item \textbf{Benchmark Distribution:} Hard (40\%) | Medium (50\%) | Easy (10\%)
\end{itemize}

\subsection{Prompt Engineering \& Schema Description}
We implemented an enhanced prompt strategy to minimize syntax errors:
\begin{enumerate}
    \item \textbf{Schema Details Generation:} Automated generation of concise schema descriptions from the Spider dataset. This enables the LLM to know details (like exact column names, data-types of columns, sample values present in columns) about the entire Database schema, so that model can take accurate and confident decisions on column selections and “where clauses” formation.
    \item \textbf{Self-Correction:} Improved system prompts that explicitly instruct the model to re-verify column names and syntax before outputting the final SQL. This ensures the model has less hallucination and has better query accuracy.
\end{enumerate}

\section{Experimental Setup}
To ensure reproducibility and a consistent basis for comparison across disparate model sizes, we standardized our model selection and evaluation framework.
\begin{itemize}
    \item \textbf{Large Models (Baselines):} We utilized the Google Gemini API for access to Gemini 2.5 Pro, Flash and Flash-Lite variants~\cite{gemini}.
    \item \textbf{Small Model:} We selected the Qwen-7B-Chat model~\cite{qwen} as our primary small language model.
    \item \textbf{Fine-Tuning Configuration:} We utilized Low-Rank Adaptation (LoRA) for fine-tuning to demonstrate parameter efficiency.
\end{itemize}

\section{The Baseline: Standard Fine-Tuning}
We compared the impact of Standard Supervised Fine-Tuning (SFT) on two distinct model classes: Large (Gemini 2.5 Family) and Small (Qwen).

\subsection{Large Model Plateau (Gemini 2.5 Experiments)}
We fine-tuned Gemini 2.5 Flash and Flash-Lite on the spider dataset. Contrary to the expectation of incremental gains, the results demonstrated that applying standard Supervised Fine-Tuning (SFT) to already highly capable models leads to performance degradation. Rather than improving upon their zero-shot baseline, the models experienced a regression in accuracy.

\textbf{Results on Full Balanced Set (600 Queries):}
\begin{itemize}
    \item \textbf{Gemini 2.5 Flash:}
    \begin{itemize}
        \item Base Model: 73.17\%
        \item Fine-Tuned: 72.50\% (Slight regression)
    \end{itemize}
    \item \textbf{Gemini 2.5 Flash-Lite:}
    \begin{itemize}
        \item Base Model: 73.00\%
        \item Fine-Tuned: 73.83\% (Marginal $<$1\% improvement)
    \end{itemize}
\end{itemize}

\paragraph{The Overfitting Risk}
When tested specifically on the top 40 complex queries, the limitations of standard fine-tuning became apparent.
\begin{itemize}
    \item Gemini 2.5 Flash (Base): 62.5\%
    \item Gemini 2.5 Pro (Base): 62.5\%
    \item Gemini 2.5 Flash (Fine-Tuned): 57.5\%
\end{itemize}

\textbf{Observation:} The drop in accuracy on complex queries indicates overfitting. 

\medskip 

\textbf{Hypothesis for Large Model Degradation:} We hypothesize that for models with massive parameter counts and extensive pre-training, standard SFT acts as a restrictive mechanism rather than an additive one. These models already possess generalized SQL generation capabilities. By fine-tuning them on standard prompt-to-SQL pairs without explicit reasoning traces, we inadvertently cause a ``representation collapse.'' The model is forced to abandon its deep, multi-step reasoning pathways in favor of shallow, dataset-specific pattern matching. Consequently, when confronted with highly complex queries, the fine-tuned large model attempts to recall a memorized structural template rather than reasoning dynamically through the schema, leading to the observed performance degradation.

\subsection{The Small Model Uplift (Qwen Experiments)}
We then applied the same standard dataset to the small-scale Qwen model. Unlike Gemini, the small model benefited from rote learning, though it still lacked reasoning depth.
\begin{itemize}
    \item Qwen Base: 36.17\%
    \item Qwen SFT (Standard): 45.33\% (+9.16\%)
\end{itemize}
While we see an improvement, 45\% accuracy is insufficient for enterprise use.

\section{The Catalyst: Chain-of-Thought (CoT) Fine-Tuning}
To push the boundaries of the small model, we hypothesized that the model lacked the reasoning process required to map natural language to structured data. We created a specialized dataset including detailed Chain-of-Thought (CoT) steps: Query Analysis, Table Selection, Column Selection, and Logic Strategy.Below are details on the dataset 

\subsection{Anatomy of the CoT Dataset}
The CoT training data explicitly modeled the step-by-step logic a human data engineer uses to write SQL. The reasoning steps included:
\begin{enumerate}
    \item \textbf{Query Analysis:} Breaking down the user's intent and identifying keywords.
    \item \textbf{Table Selection:} Analyzing the provided schema to select only the relevant tables.
    \item \textbf{Column Selection:} Filtering the chosen tables for the exact columns needed to satisfy the query.
    \item \textbf{Logic \& Join Strategy:} Determining the primary/foreign key relationships needed for joins.
    \item \textbf{Self-Validation:} A final internal check to ensure the generated SQL directly answers the user's intent without hallucinating columns.
\end{enumerate}

This is how one example of sample training data with COT looks like:
\vspace{1em}
\begin{lstlisting}[language=json, caption=Full Sample CoT Training Data Example, breaklines=true, basicstyle=\scriptsize\ttfamily, frame=single]
{
  "messages": [
    {
      "role": "system",
      "content": "You are a powerful text-to-SQL model. Your role is to answer user questions by generating SQL queries against a given database schema. First, provide a step-by-step chain of thought that explains your reasoning, and then provide the final SQL query in a markdown code block."
    },
    {
      "role": "user",
      "content": "DATABASE SCHEMA:
Table: Customers
customer_id INTEGER PRIMARY KEY
customer_first_name VARCHAR(50)
customer_middle_initial VARCHAR(1)
customer_last_name VARCHAR(50)
gender VARCHAR(1)
email_address VARCHAR(255)
login_name VARCHAR(80)
login_password VARCHAR(20)
phone_number VARCHAR(255)
town_city VARCHAR(50)
state_county_province VARCHAR(50)
country VARCHAR(50)

Table: Orders
order_id INTEGER PRIMARY KEY
customer_id INTEGER NOT NULL
date_order_placed DATETIME NOT NULL
order_details VARCHAR(255)

Table: Invoices
invoice_number INTEGER PRIMARY KEY
order_id INTEGER NOT NULL
invoice_date DATETIME

Table: Accounts
account_id INTEGER PRIMARY KEY
customer_id INTEGER NOT NULL
date_account_opened DATETIME
account_name VARCHAR(50)
other_account_details VARCHAR(255)

Table: Product_Categories
production_type_code VARCHAR(15) PRIMARY KEY
product_type_description VARCHAR(80)
vat_rating DECIMAL(19,4)

Table: Products
product_id INTEGER PRIMARY KEY
parent_product_id INTEGER
production_type_code VARCHAR(15) NOT NULL
unit_price DECIMAL(19,4)
product_name VARCHAR(80)
product_color VARCHAR(20)
product_size VARCHAR(20)

Table: Financial_Transactions
transaction_id INTEGER NOT NULL
account_id INTEGER NOT NULL
invoice_number INTEGER
transaction_type VARCHAR(15) NOT NULL
transaction_date DATETIME
transaction_amount DECIMAL(19,4)
transaction_comment VARCHAR(255)
other_transaction_details VARCHAR(255)

Table: Order_Items
order_item_id INTEGER PRIMARY KEY
order_id INTEGER NOT NULL
product_id INTEGER NOT NULL
product_quantity VARCHAR(50)
other_order_item_details VARCHAR(255)

Table: Invoice_Line_Items
order_item_id INTEGER NOT NULL
invoice_number INTEGER NOT NULL
product_id INTEGER NOT NULL
product_title VARCHAR(80)
product_quantity VARCHAR(50)
product_price DECIMAL(19,4)
derived_product_cost DECIMAL(19,4)
derived_vat_payable DECIMAL(19,4)
derived_total_cost DECIMAL(19,4)

Question: What are the ids, date opened, name, and other details for all accounts?"
    },
    {
      "role": "assistant",
      "content": "To answer the question \"What are the ids, date opened, name, and other details for all accounts?\", we need to perform the following steps:

1. Identify the target table: The question explicitly asks for information about \"accounts\". Looking at the schema, the Accounts table is the most relevant table as it stores account-related data.

2. Identify the required columns: The question asks for \"ids, date opened, name, and other details\".
    * \"ids\" corresponds to the account_id column in the Accounts table.
    * \"date opened\" corresponds to the date_account_opened column in the Accounts table.
    * \"name\" corresponds to the account_name column in the Accounts table.
    * \"other details\" corresponds to the other_account_details column in the Accounts table.

3. Determine if filtering is needed: The question asks for \"all accounts\", which implies no specific filtering conditions are required.

4. Determine if joins are needed: All the requested information is available directly within the Accounts table, so no joins with other tables are necessary.

5. Determine if aggregation or ordering is needed: The question does not ask for any summary statistics (like counts, sums, averages) or a specific order for the results, so no aggregation or ordering clauses are needed.

Therefore, the query will simply select the identified columns from the Accounts table.

```sql
SELECT account_id, date_account_opened, account_name, other_account_details FROM Accounts
```"
    }
  ]
}
\end{lstlisting}

\subsection{Results: A Paradigm Shift}
When the small Qwen model was fine-tuned on this CoT dataset, it significantly outperformed the standard fine-tuning approach.

\begin{itemize}
    \item Qwen Base: 36.17\%
    \item Qwen SFT (No CoT): 45.33\%
    \item \textbf{Qwen CoT: 54.50\%}
\end{itemize}

\textit{Analysis:} By  explicitly training the small model to generate intermediate reasoning traces rather than mapping inputs directly to final outputs, we achieved an 18.33\% absolute increase over the zero-shot baseline. Notably, the Qwen CoT model (54.50\%) demonstrates that transferring these multi-step reasoning pathways enables a compute-efficient architecture to significantly narrow the performance gap with larger models, at a fraction of the inference cost.

\section{Detailed Experimental Data}
The following table summarizes the performance across all tested architectures and datasets.

\begin{table}[H]
\centering
\caption{Comparative Accuracy \& Configuration}
\resizebox{\textwidth}{!}{%
\begin{tabular}{@{}llllll@{}}
\toprule
\textbf{Exp ID} & \textbf{Model} & \textbf{Training Method} & \textbf{Dataset Size} & \textbf{Hyperparameters} & \textbf{Exec Acc (600 Q)} \\ \midrule
G-Flash-Base & Gemini 2.5 Flash & Zero-shot & N/A & N/A & 73.17\% \\
G-Flash-FT & Gemini 2.5 Flash & Standard SFT & Full Spider & Ep: 3, LRM: 1, Adpt: 16 & 72.50\% \\
G-Lite-Base & Gemini 2.5 F-Lite & Zero-shot & N/A & N/A & 73.00\% \\
G-Lite-FT & Gemini 2.5 F-Lite & Standard SFT & Full Spider & Ep: 3, LRM: 5e-5, Adpt: 16 & 73.83\% \\
Qwen-Base & Qwen (7b) & Zero-shot & N/A & N/A & 36.17\% \\
Qwen-SFT & Qwen (7b) & Standard SFT & Full Spider & Ep: 3, LRM: 2e-5, Adpt: 16 & 45.33\% \\
\textbf{Qwen-CoT} & \textbf{Qwen (7b)} & \textbf{CoT Fine-Tune} & \textbf{Spider + CoT} & \textbf{Ep: 5, LRM: 2e-5, Adpt: 32} & \textbf{54.50\%} \\ \bottomrule
\end{tabular}%
}
\footnotesize{\\ *LRM - learning rate multiplier, Ep - Epochs Note: ``Execution Failed'' rates were highest in Qwen Base (223 failures), drastically reduced in CoT (106 failures), indicating significantly improved syntax generation.}
\end{table}

\begin{figure}[H]
    \centering
    \includegraphics[width=0.8\textwidth]{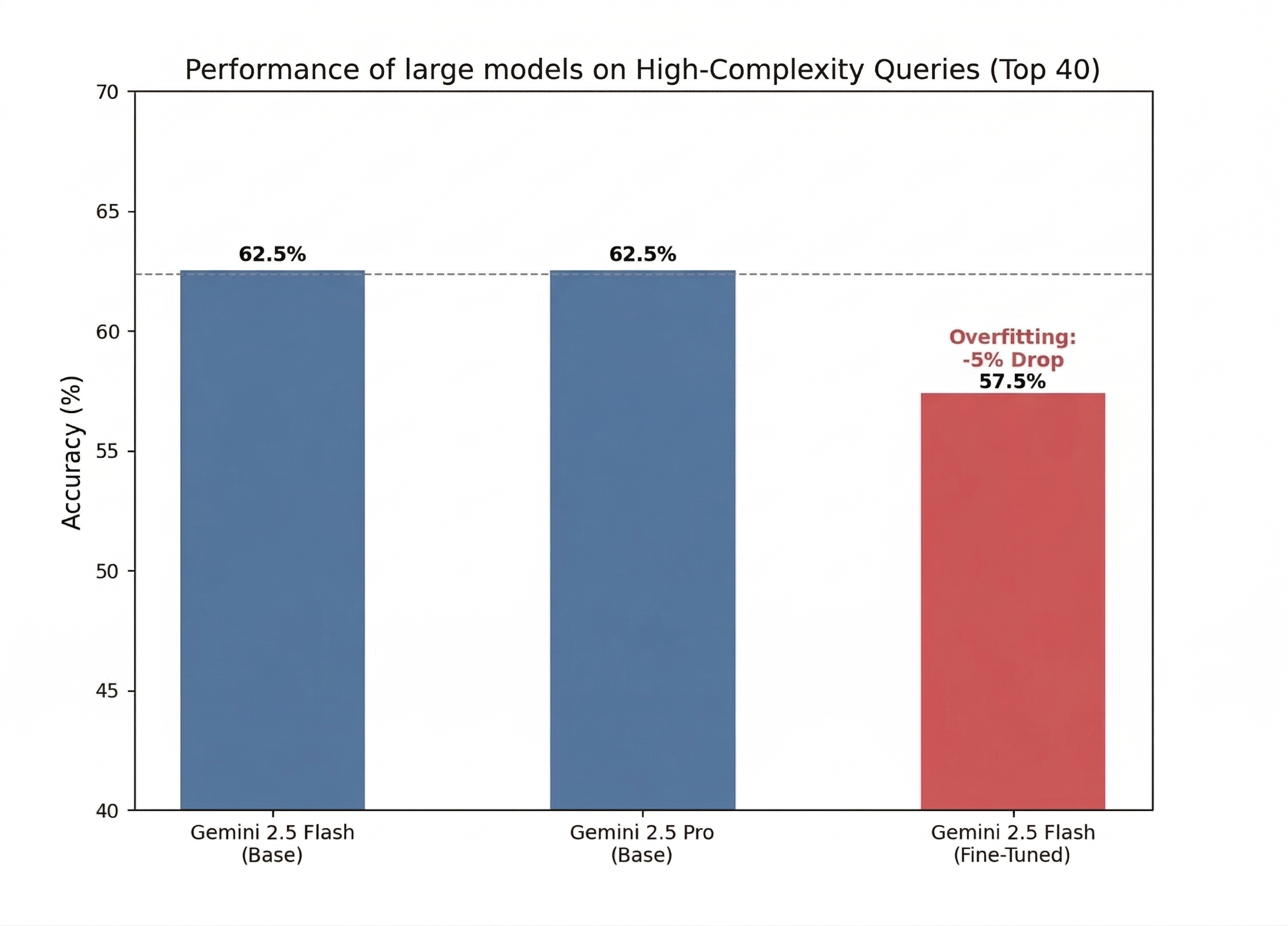}
    \caption{Large Model Performance.}
    \label{fig:large_models}
\end{figure}

\begin{figure}[H]
    \centering
    \includegraphics[width=0.8\textwidth]{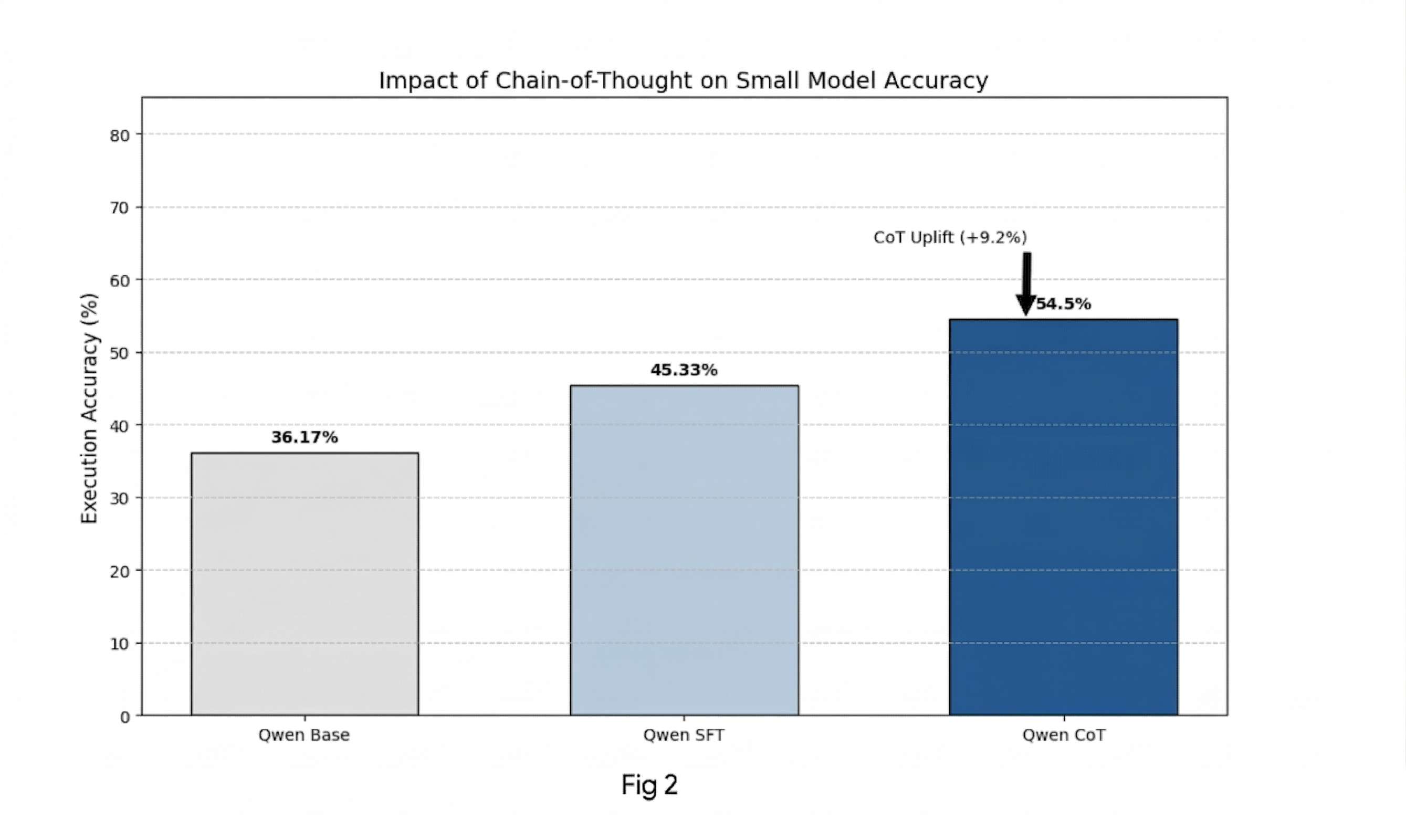}
    \caption{Small Model Performance.}
    \label{fig:small_models}
\end{figure}

\begin{figure}[H]
    \centering
    \includegraphics[width=0.8\textwidth]{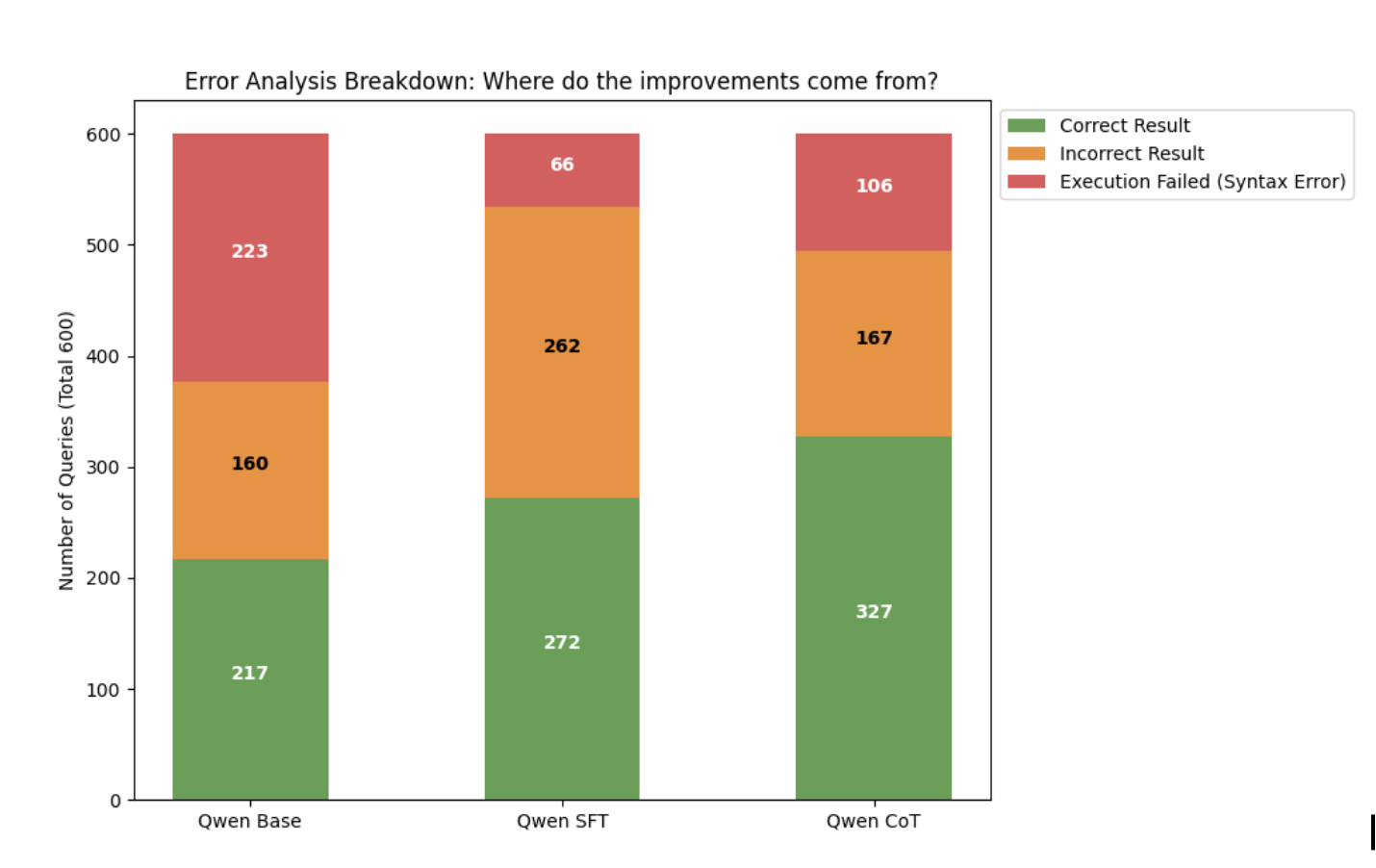}
    \caption{Error Analysis Breakdown.}
    \label{fig:error_img}
\end{figure}

\section{Discussion and Enterprise Implications}
The data confirms a distinct split in strategy based on model size:
\begin{itemize}
    \item \textbf{For Large Models (Gemini 2.5):} Fine-tuning is largely unnecessary and potentially harmful (overfitting) for general SQL tasks. Prompt engineering and RAG are more effective levers.
    \item \textbf{For Small Models (Qwen):} CoT Fine-tuning is essential. It bridges the reasoning gap, allowing a model that originally failed 64\% of queries to answer the majority correctly.
\end{itemize}

\paragraph{Implications:}
\begin{itemize}
    \item \textbf{Massive Compute Cost Reduction:} Achieving $\approx$55\% accuracy on a small model allows for cost-effective edge deployment.
    \item \textbf{Latency:} Small models generate tokens faster, offsetting the extra tokens required for CoT reasoning steps.
    \item Small models can achieve production-viable performance on fine-tuning with COT data.
\end{itemize}
\section{Limitations}
While our Chain-of-Thought fine-tuning approach significantly narrows the performance gap between small and large models, several limitations still remain. 

First, our evaluation is restricted to the Spider dataset's development split. While Spider is a robust benchmark, real-world enterprise databases often contain messy, unnormalized schemas and ambiguous column naming conventions not fully represented in academic datasets. Second, the "token overhead" of CoT is non-negligible. Although the Qwen model is small, generating a verbose explanation before the SQL query increases output latency compared to direct SQL generation. Future work will investigate "internalized" reasoning, where the model is trained to reason in latent space rather than generating explicit text.

\section{Conclusion}
Fine-tuning LLMs for NL2SQL tasks requires a nuanced approach based on model scale. While large models do not significantly benefit from standard fine-tuning, smaller models show immense potential when trained properly.

Our research demonstrates that high-quality, reasoning-dense data (Chain-of-Thought) is the key to unlocking large-model capabilities in small-model architectures. For NL2SQL, teaching a model to reason through table selection, column matching, and self-validation is vastly superior to rote memorization of query pairs.

\end{document}